# VPI-MLOGS: A WEB-BASED MACHINE LEARNING SOLUTION FOR APPLICATIONS IN PETROPHYSICS

**Nguyen Anh Tuan**
Vietnam Petroleum Institute
Email: tuan.a.nguyen@vpi.pvn.vn

## Summary

Machine learning is an important part of the data science field. In petrophysics, machine learning algorithms and applications have been widely approached. In this context, Vietnam Petroleum Institute (VPI) has researched and deployed several effective prediction models, namely missing log prediction, fracture zone and fracture density forecast, etc. As one of our solutions, VPI-MLogs is a web-based deployment platform which integrates data preprocessing, exploratory data analysis, visualisation and model execution. Using the most popular data analysis programming language, Python, this approach gives users a powerful tool to deal with the petrophysical logs section. The solution helps to narrow the gap between common knowledge and petrophysics insights. This article will focus on the web-based application which integrates many solutions to grasp petrophysical data.

## 1. Introduction

Understanding data is a crucial step in any aspect of technological fields and research domains. In data science, clearly and precisely understanding data always requires time. In the petroleum field, petrophysics data has several unique features that require users to have not only domain knowledge but also specialised software to deal with data problems.

The most notable programming languages (such as Python) give developers tools to address issues and validate data without any special softwares or payments. In addition, some valuable functions could be designed to fit the user's machine learning requirements such as data processing, data cleaning, exploratory data analysis and model deployment.

The dashboard is basically fulfilled by charts, model results, and data insights. For example, Power BI and Tableau take a lot of advantages by their powerful organised abilities. However, because of their limited modification, several innovative ideas cannot be presented. Alternatively, many Python libraries appeared to support presentation and graphic user interface functions. Streamlit.io is one of these answers, combined with interactive visualisation by Altair library helping improve display features and data exploration.

In the end, a solution integrating interactive visuals and web applications has completely erected to deal with petrophysical log data which include several steps from data preprocessing (LAS files loading and re-organising, EDA, outliers removal, etc.) to model deployment (missing log forecast or fracture prediction). A web-based application is also more friendly than rigid coding lines.

## 2. Recent work and new approach

Traditionally, most of petrophysical tasks require custom software such as Petrel, Techlog (Schlumberger), IP Interactive Petrophysics (Lloyd's Register)...

During log interpretation, interactive function is performed beside advanced operations to provide information for exploration progress. On the other hand, recently, machine learning algorithms have become more and more popular and embedded in almost all industrial sectors. However, updating the latest technology always faces many restrictions, especially in financial aspect. From the user's perspective, VPI's team has researched and experienced applications of machine learning to address missing log data or erect fracture predictive models.

In operation perspective, professional software runs locally in user's devices. It always requires a computer with high performance, and in some cases, it needs a workstation. This traditional approach retains several limitations such as high cost or immobility.

To solve these issues, the web-based application is considered. The new approach focuses on execution velocity and convenience with many advantages, namely the ability of implementing on medium performance computation, online availability, easy to access and ease-of-use. Following to the solution, users can upload their log data to the application host then predictive models are performed to return results back to users.

## 3. Research method

Python has grown to be one of the most popular programming languages in the world and is widely adopted in the data science community. Python contains a wide range of tools such as Pandas for data manipulation and analysis, Matplotlib for data visualisation, and Scikit-learn for machine learning, all aimed towards simplifying different stages of the data science pipeline.

Python supports a lot of visualisation libraries that allow users to generate data insights. There are prominent libraries with unique features such as matplotlib, seaborn, plotly, etc.. Recently, the performance has been further enhanced with the emergence of interactive visualisation tools.

To adapt for a web-based approach, several libraries in Python have been used namely Pandas, NumPy, Matplotlib, Altair, Streamlit, Vega-Lite.

### *3.1. Interactive visuals*

In recent petrophysical log interpretation process, the main activities are handled by interactive windows such as histogram, cross-plot charts, curve view, etc. Therefore, interaction function always plays an important role. Instead of using merely traditional visual libraries (matplotlib, seaborn, etc.), the new approach focuses on a novel visual technique which is optimised to deal with hundreds of thousands dataset instances, and the most important, it possesses supreme interaction functions.

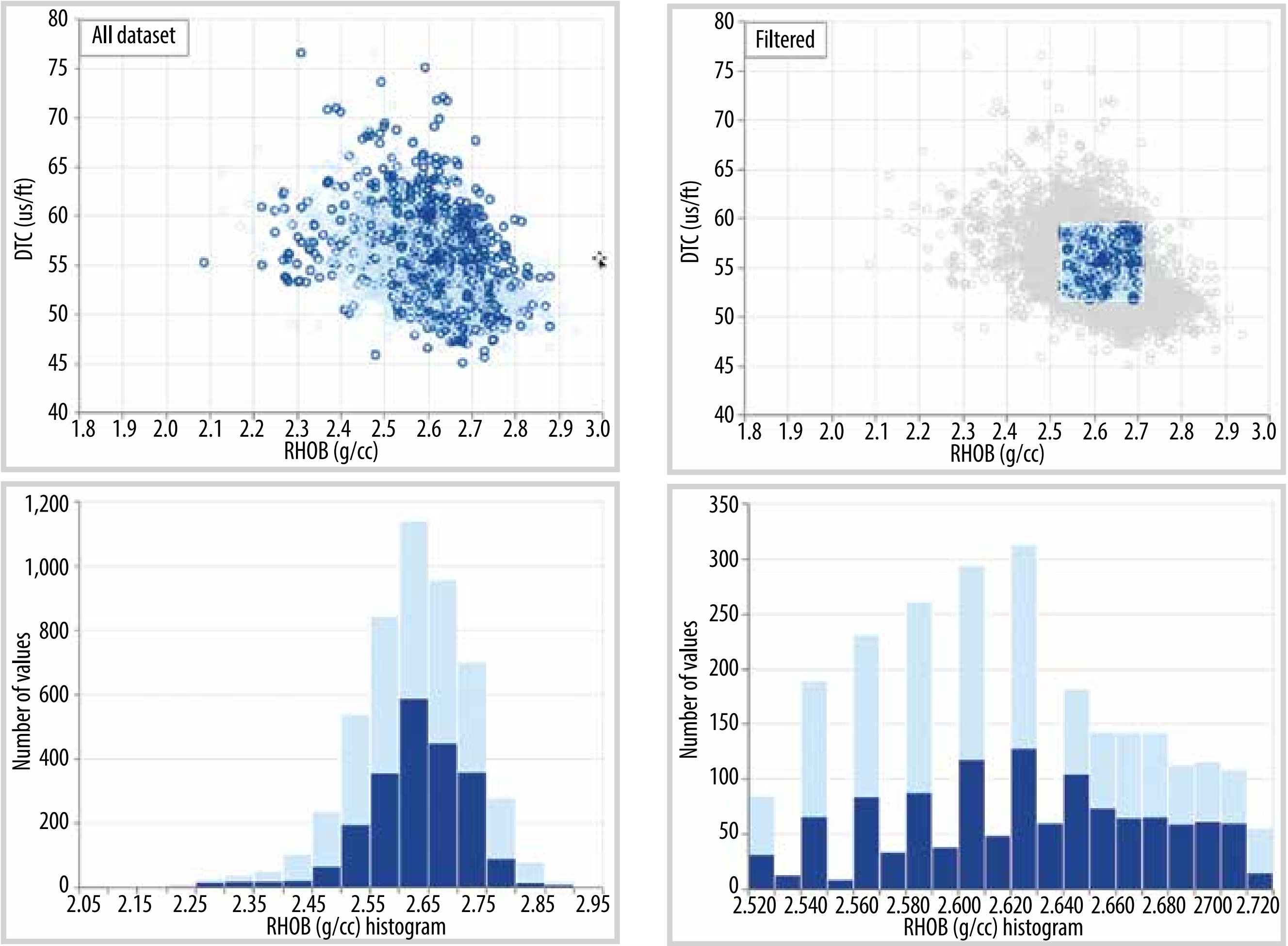

***Figure 1.*** *Scatter plot and histogram chart interact with user selection.*

Altair is a visualisation Python library based on the Vega-Lite grammar, which allows a wide range of statistical visualisations to be expressed using a small number of grammar primitives. Vega-Lite implements a view composition algebra in conjunction with a novel grammar of interactions that enables users to specify interactive charts in a few lines of code. Vega-Lite is declarative; visualisations are specified using JSON data that follows the Vega-Lite JSON schema [1].

Altair allows users to directly interact with charts and connect to different visualisations. In Figure 1, a cross-plot between DTC and RHOB is represented simultaneously with the RHOB histogram chart. By interaction from the cross-plot view, selected points are immediately filtered in the histogram charts. It brings a convenient approach to understand petrophysical data as well as interpret initial mutual relation between logs.

### *3.2. Web-based framework to deploy our machine learning model*

Beyond a visual dashboard, the model deployment solution should be considered. PowerBI or Tableau seems to be limited. The appearance of Streamlit in 2020 swiftly received great attention thanks to its many advantages in terms of speed, readability, ease-of-use and the ability of operating predictive models on the web-base.

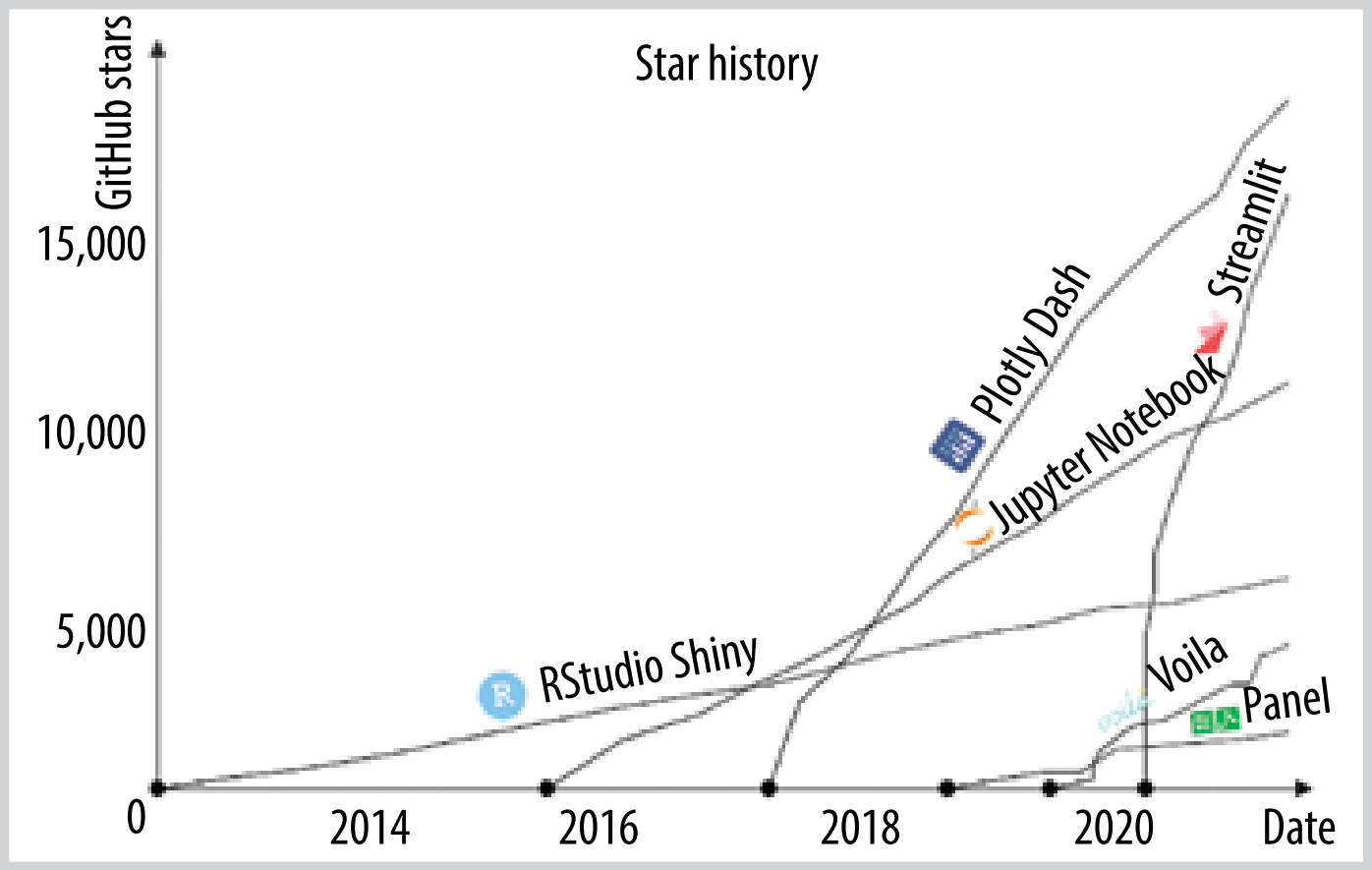

***Figure 2.*** *Streamlit has surged in popularity in recent years.*

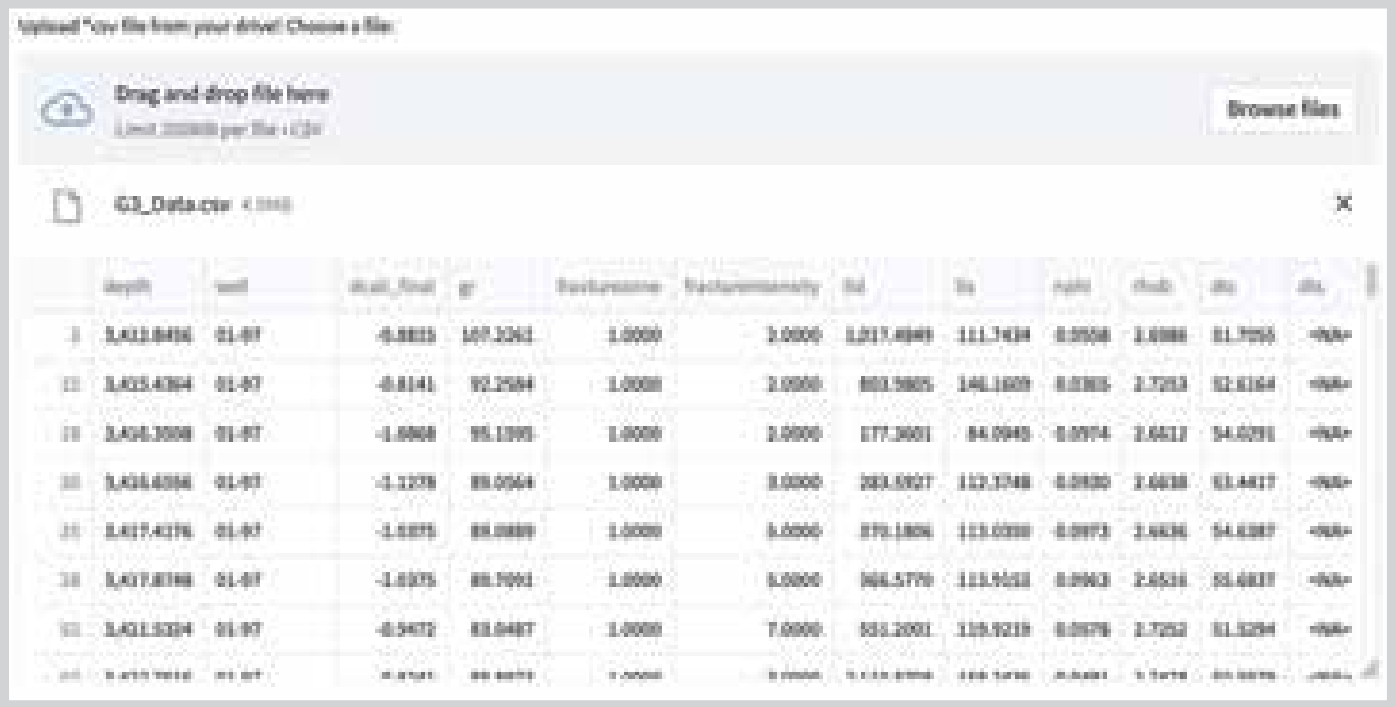

***Figure 3.*** *Application interface and loading data section.*

Generally, Streamlit is an open-source Python library that is used to build powerful, custom web applications for data science and machine learning. Streamlit is compatible with several major libraries and frameworks such as Latex, OpenCV, Vega-Lite, seaborn, PyTorch, NumPy, Altair, and more. Streamlit is also popular and used among big industry leaders, such as Uber and Google X.

Besides, Streamlit has a wide range of UI components. It covers almost every common UI component such as checkbox, slider, a collapsible sidebar, radio buttons, file upload, progress bar, etc. Moreover, these components are very easy to use.

Streamlit has made it thoroughly simple to create interfaces, display text, visualise data, render widgets, and manage a web application from inception to deployment with its convenient and highly intuitive application programming interface [2].

## 4. VPI-MLogs for petrophysics

The application named VPI-MLogs includes full steps of a machine learning project to deal with petrophysical log problems. It can be summarised in 4 main stages: Data collection, data cleaning/processing, EDA, Model&Prediction. The solution is deployed on a web-based platform thus avoiding the requirements of specialised software and technical expertise.

### *4.1. Data collection*

Every petrophysical log data is stored as Log ASCII Standard (LAS) format with tabular structure. In Python, Lasio library allows users to access information directly from LAS files and transfer it to tabular data (pandas DataFrame). All calculations and modifications are made conveniently after this conversion.

In the first step, data in the LAS extension can be collected and loaded to the dashboard. The system automatically converts it to pandas DataFrame. Several functions are also provided for users' modification, namely curves name changing, setting the limited values, saving selected curves, merging multiple LAS files to CSV file…

Then, a preprocessed database has been formed and ready for the upcoming stages. A download button allows users to save the revised file to their storage.

### *4.2. Data cleaning/processing*

During model preparation, it is important to clean the data sample to ensure that the observations best represent the problem. Outliers are unusual values in the dataset, and in general, machine learning modelling and modelling processes can be improved by understanding and even removing these values. In petrophysical logs, outliers can be resulted from many reasons: measurement errors, drilling fluid impact, bore well collapse, etc.

Even with a thorough understanding of the data, outliers can be hard to define. Great care should be taken

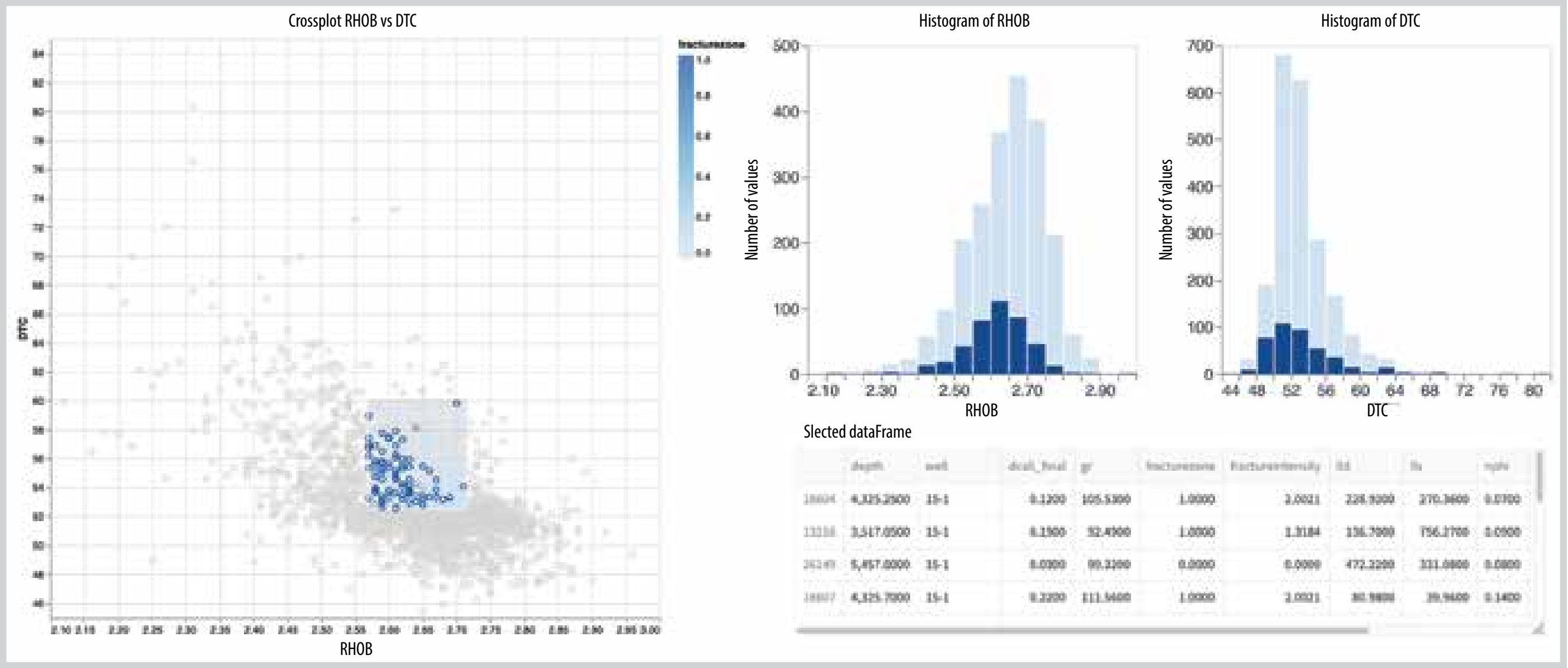

***Figure 4.*** *Streamlit selection integrated with cross-plot and histogram chart to highlight the outliers. The outlier data points showed by selection (right-bottom).*

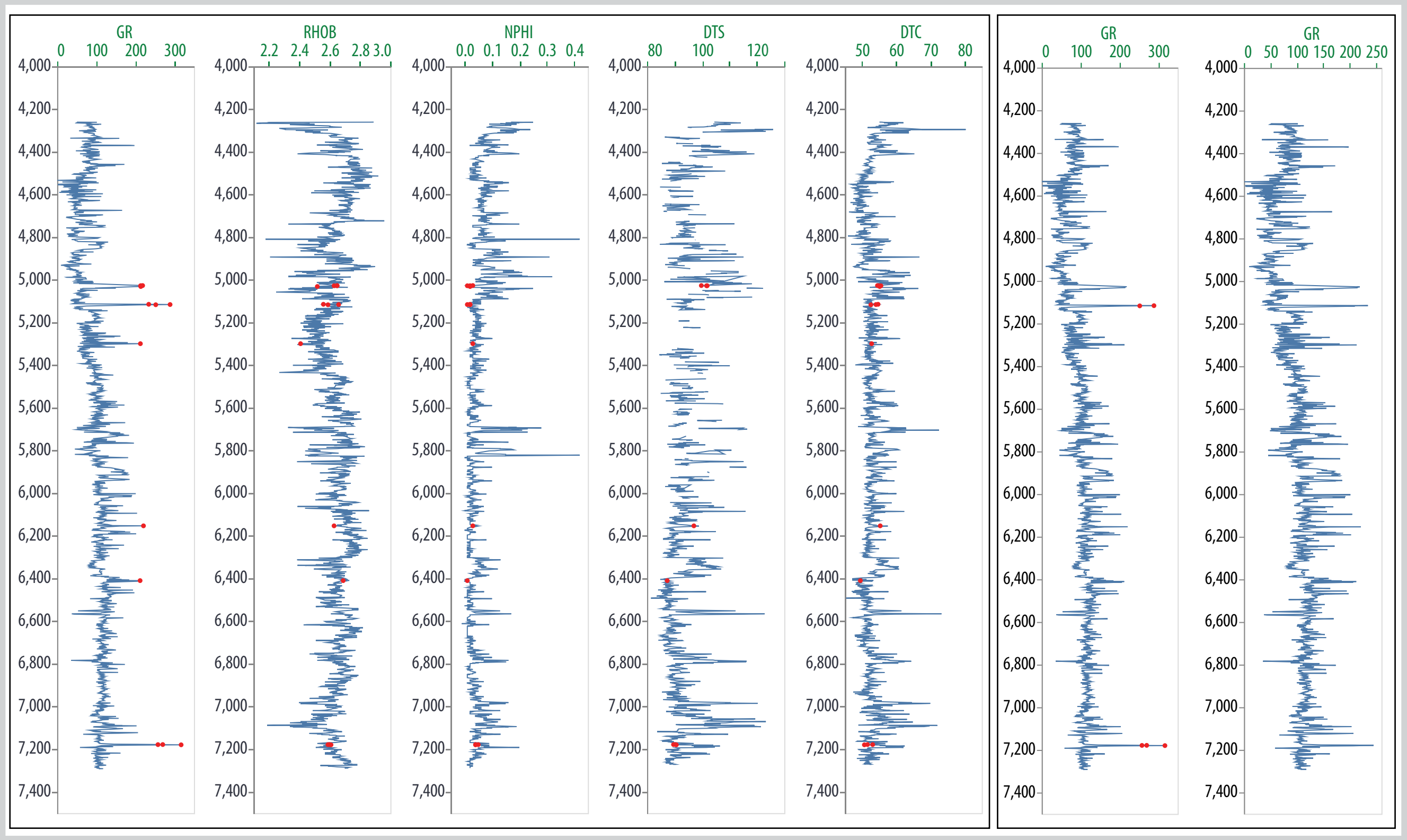

(a) (b)

***Figure 5.*** *Curves and the highlight of selected point in log view (a). Outliers removing (b).*

not to remove or change values hastily, especially if the sample size is small [2].

On the web-based solution, many types of functions such as histogram chart, cross-plot 2 curves, logs view provide a basic tool to detect outliers. By user's selection of wells and curves to plot, they can proactively interact with data and deal with the skeptical points.

Combined with cross-plot, selection is an important tool to detect suspicious outliers (Figure 4). Simultaneously, the skeptical points are indicated in the dataset and highlighted on the curve view charts. Expert users using analytical techniques will decide whether to remove the outlier or keep it as good data points. The result can be saved to the local disk by a download button.

### 4.3. Exploratory data analysis

Exploratory data analysis is the stage where we actually start to understand the message contained in the data. EDA examines what data can tell us before actually going through formal modeling or hypothesis formulation. It should be noted that several types of data transformation techniques might be required during the process of exploration [3].

Several visuals have been equipped and integrated to support the EDA process:

- Scatter graphs: Scatter plots are used when we need to show the relationship between two variables. These plots are powerful tools for visualisation, despite their simplicity.

- Histogram: Histogram plots are used to depict the distribution of any continuous variable. These types of plots are very popular in statistical analysis.

- Bar charts: Bar charts are frequently used to distinguish objects between distinct collections to track variations over time. Bars can be drawn horizontally or vertically to represent categorical variables.

- Box plot: A type of descriptive statistics chart, visually shows the distribution of numerical data and skewness through displaying the data quartiles (or percentiles) and averages.

- Pair plot: A simple way to visualise relationships

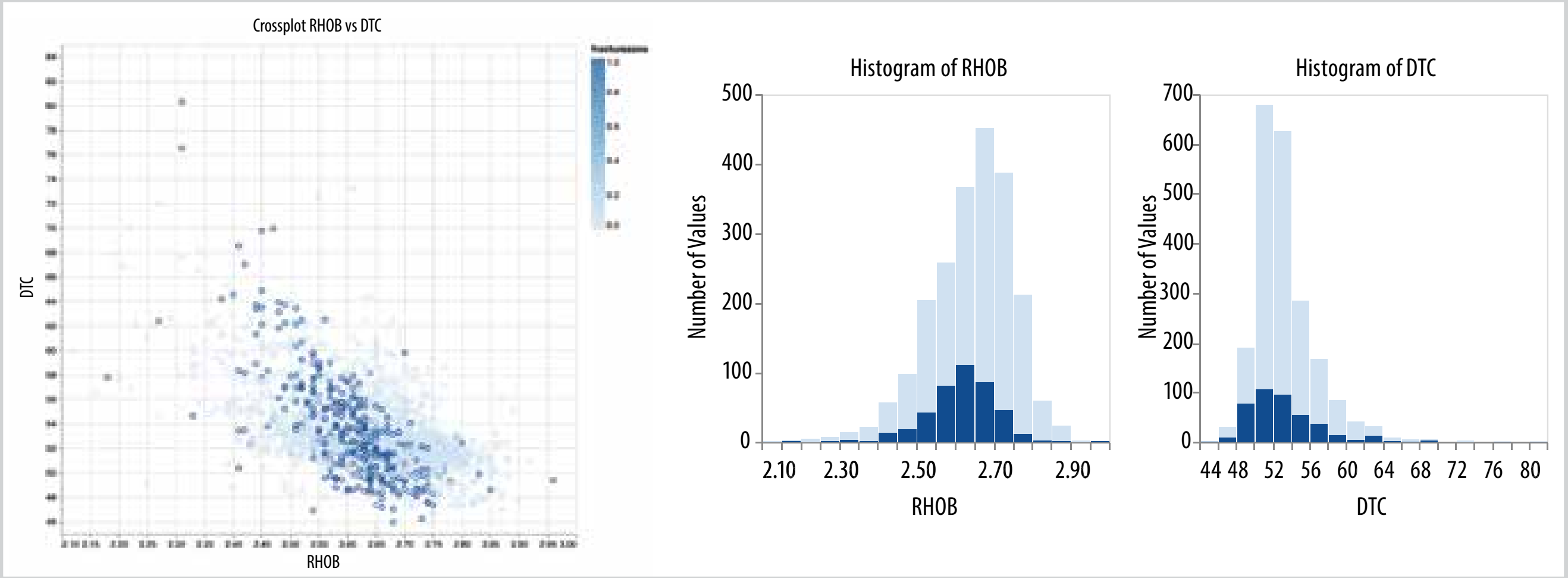

**Figure 6.** *Scatter and histogram charts.*

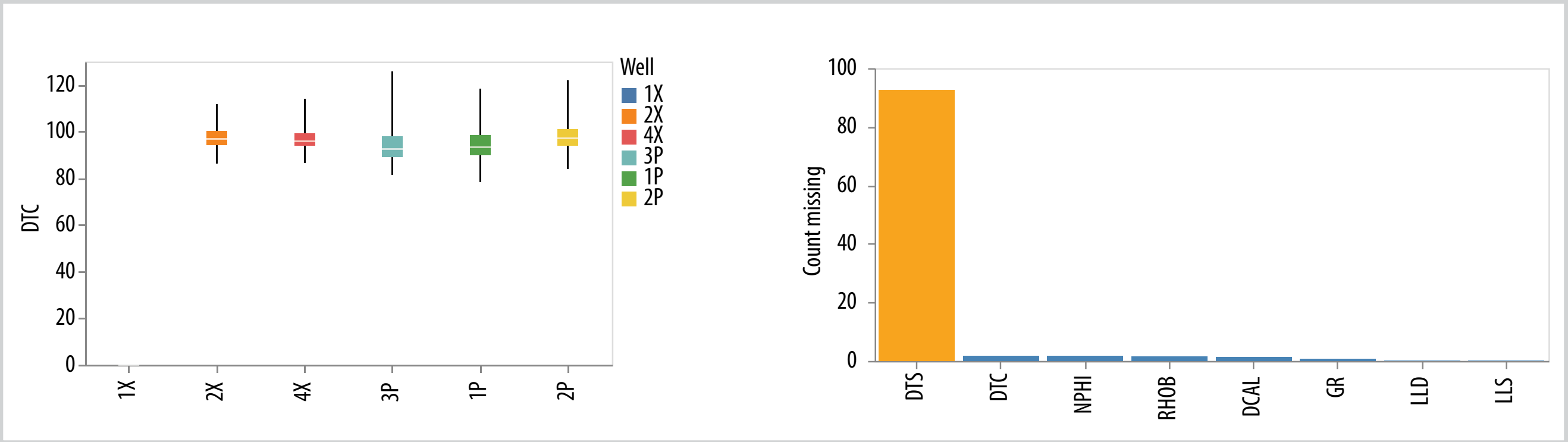

**Figure 7.** *Bar chart and box plot.*

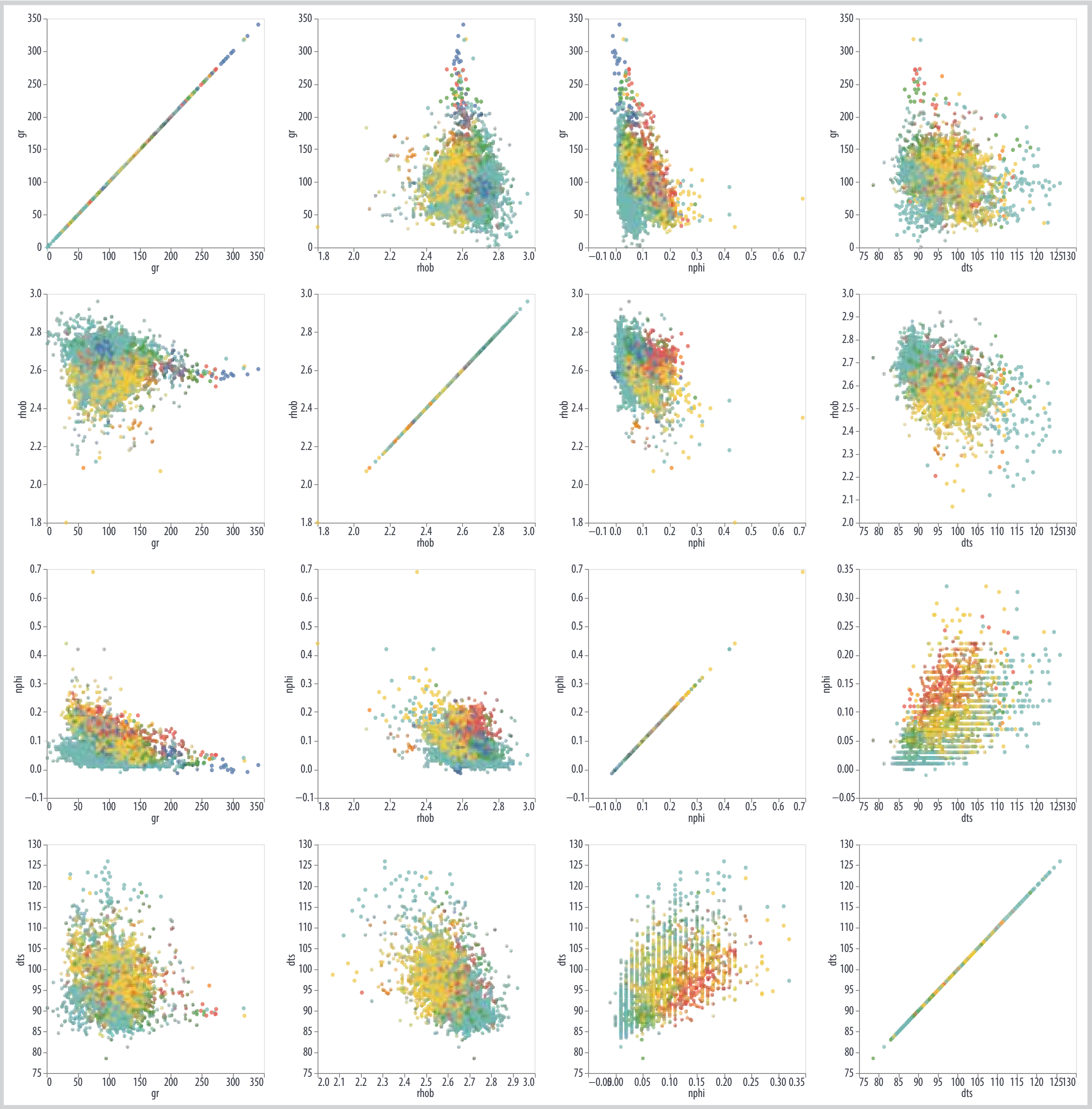

***Figure 8.*** *Pair plot shows the duo relationship among logs.*

between each variable. It produces a matrix of relationships between each variable in the data for an instant data examination.

- Correlation heatmap: A type of plot that visualises the strength of relationships between numerical variables. Correlation plots are used to understand which variables are related to each other and the strength of this relationship.

### *4.4. Model deployment and prediction*

Model deployment is the process of putting machine learning models into production. This makes the model's predictions available to users, developers or systems.

Streamlit is an alternative to Flask for deploying the machine learning model as a web service. The biggest advantage of using Streamlit is that it allows users to use HTML code within the application Python file. It doesn't essentially require separate templates and CSS formatting for the front-end UI [4].

In this section, a fitting predictive model is added under the code. Following that, the cleaned data containing features, which can be loaded by users, will be used as input of the model. By click the prediction button on the interface, the prediction process can be operated. The output is a dataset with predicted values. In addition, visual curves will appear.

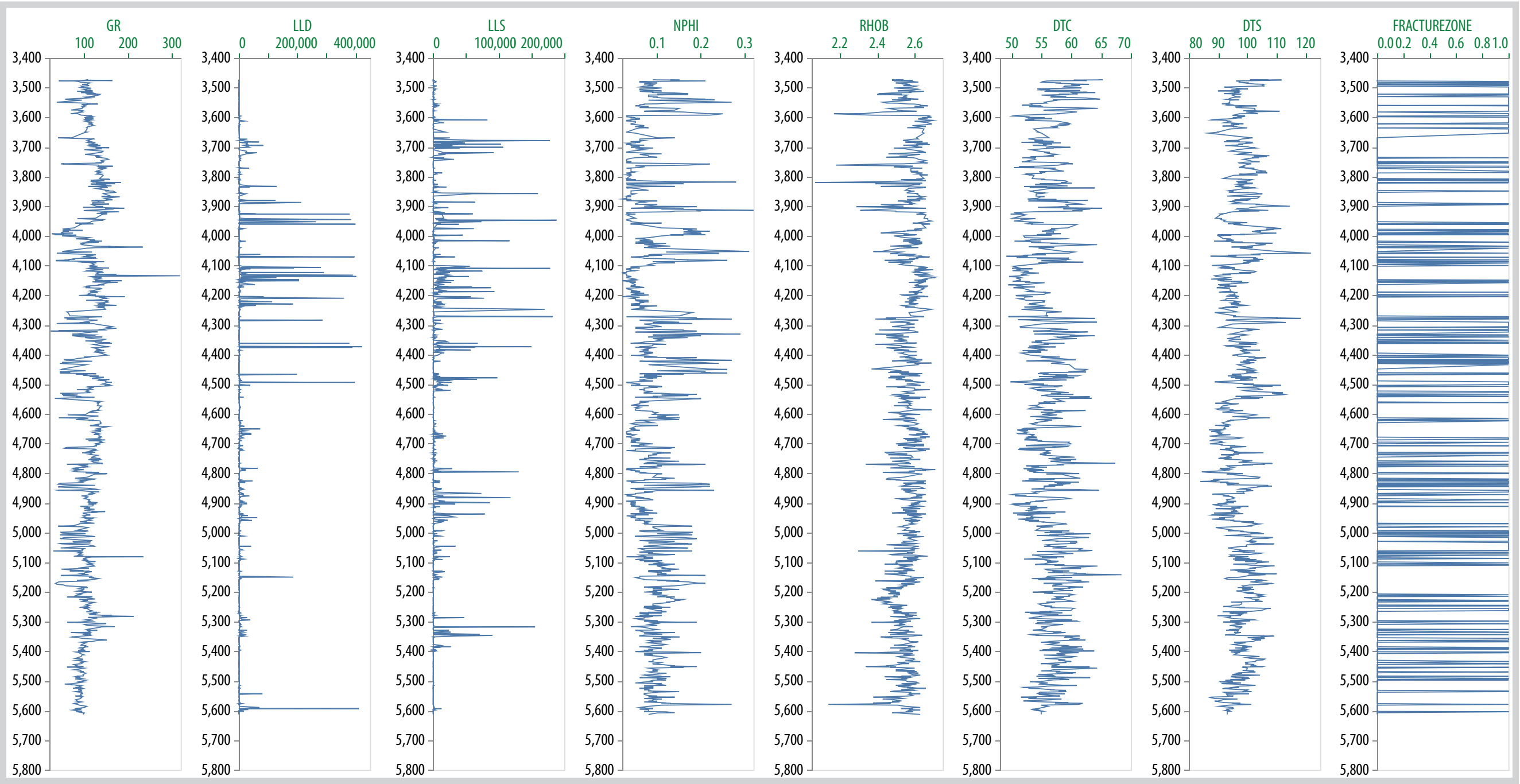

***Figure 9.*** *Curves view with predicted values.*

In Figure 9, data uploaded from users include GR, LLD, LLS, NPHI, RHOB, DTC and DTS used as features which have been put to the fracture predictive model. Besides, the prediction result depicted next to features graphs. Through this visualisation, users can evaluate the predicted value by cross-checking with other curves concurrently.

The results can be downloaded and saved as petrophysical logs (LAS) or CSV file.

## 5. Conclusion and future outlook

The main objective of this application is to provide a solution for petrophysical log visualisation, modification, and predictive model deployment. Python and several libraries are used to perform the functions. Altair has been used as the main tool of observation and selection. Furthermore, a web-based system has been chosen as a fast and friendly method of model deployment. In which, Streamlit stands out with the advantages of simplicity, readability. Eventually, the whole solution covers from data loading, curves modification, outliers removal and model prediction.

In upcoming stages, training progress will be included in the VPI-MLogs final solution. Then, users can use their data as training input. VPI-MLogs will allow users to change the hyperparameters and select algorithms to optimise their model. In the end, users can entirely modify their data, build their model, and finally make their prediction.